\documentclass[10pt,twocolumn,letterpaper]{article}

\usepackage{iccv}
\usepackage{times}
\usepackage{epsfig}
\usepackage{graphicx}
\usepackage{amsmath}
\usepackage{amssymb}

\newcommand{\Tref}[1]{Table~\ref{#1}}
\newcommand{\Eref}[1]{Eq.~(\ref{#1})}
\newcommand{\Fref}[1]{Fig.~\ref{#1}}
\newcommand{\Sref}[1]{Sec.~\ref{#1}}

\usepackage[breaklinks=true,bookmarks=false]{hyperref}

\iccvfinalcopy 

\def\iccvPaperID{1943} 

\ificcvfinal\fi
\begin{document}

\title{AttentionNet: Aggregating Weak Directions for Accurate Object Detection}

\author{Donggeun Yoo\\
{\small dgyoo@rcv.kaist.ac.kr}\\
{\normalsize KAIST}
\and
\hspace{-2mm}Sunggyun Park\\
\hspace{-2mm}{\small sunggyun@kaist.ac.kr}\\
\hspace{-2mm}{\normalsize KAIST}
\and
\hspace{-2mm}Joon-Young Lee\thanks{This work was done when he was in KAIST. He is currently working in Adobe Research.}\\
\hspace{-2mm}{\small jylee@rcv.kaist.ac.kr}\\
\hspace{-2mm}{\normalsize KAIST}
\and
\hspace{-2mm}Anthony S. Paek\\
\hspace{-2mm}{\small apaek@lunit.io}\\
\hspace{-2mm}{\normalsize Lunit Inc.}
\and
\hspace{-2mm}In So Kweon\\
\hspace{-2mm}{\small iskweon@kaist.ac.kr}\\
\hspace{-2mm}{\normalsize KAIST}
}

\maketitle

\begin{abstract}
We present a novel detection method using a deep convolutional neural network (CNN), named AttentionNet. We cast an object detection problem as an iterative classification problem, which is the most suitable form of a CNN. AttentionNet provides quantized weak directions pointing a target object and the ensemble of iterative predictions from AttentionNet converges to an accurate object boundary box. Since AttentionNet is a unified network for object detection, it detects objects without any separated models from the  object proposal to the post bounding-box regression. We evaluate AttentionNet by a human detection task and achieve the state-of-the-art performance of 65\% (AP) on PASCAL VOC 2007/2012 with an 8-layered architecture only.

\end{abstract}

\section{Introduction}
\begin{figure}[t]
\begin{center}
\includegraphics[width=1\linewidth]{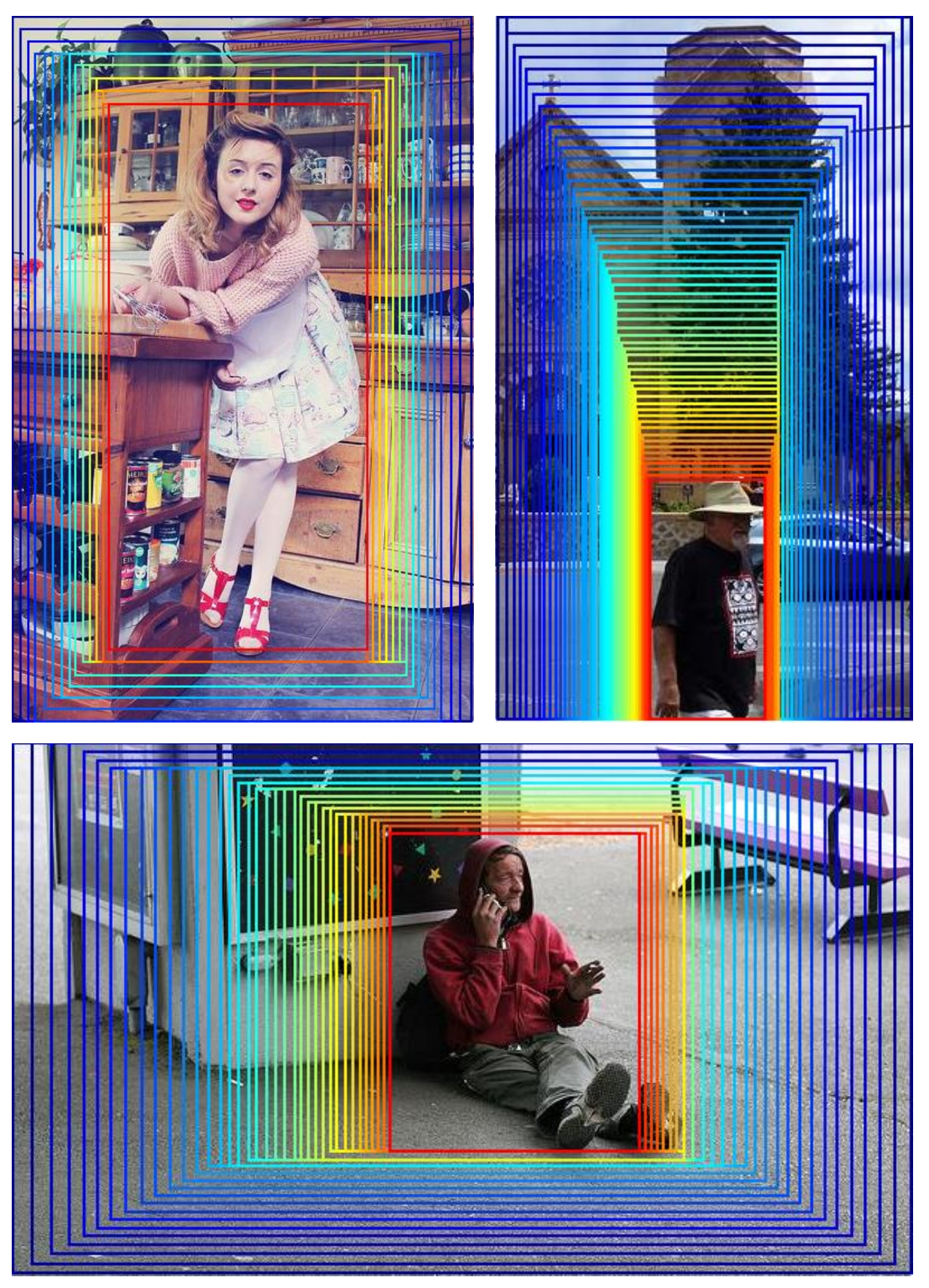}
\end{center}
   \caption{Real detection examples of our detection framework. Starting from an image boundary (dark blue bounding box), our detection system iteratively narrows the bounding box down to a final human location (red bounding box).}
\label{FIG_TEASER}
\end{figure}

After the recent advance \cite{ALEX} of deep convolutional neural network (CNN) \cite{LECUN}, CNN based object classification methods in computer vision has reached human-level performance on ILSVRC classification \cite{ILSVRC}: top-5 error of 4.94\% \cite{PRELU}, 6.67\%  \cite{GOOGLENET}, 6.8\% \cite{VERYDEEP}, and 5.1\% for human \cite{ILSVRC}. These successful methods, however, have limited range of application since most of the images are not object-centered. Thus, current research focus in visual recognition is moving towards richer image understanding such as object detection or pixel-level object segmentation. Our focus lies in the object detection problem.

Although many researchers had proposed various CNN-based techniques for object detection \cite{MASKCNN, OVERFEAT, RCNN, DEEPMULTIBOX}, it is still challenge to estimate object bounding boxes beyond the object existence. Even the most successful framework so far, Region-CNN \cite{RCNN} reported top scores \cite{DEEPIDNET, VERYDEEP, GOOGLENET} in ILSVRC'14, but it is relatively far from human-level accuracy compared to the classification. One major limitation of this framework is that the proposal quality highly affects the detection performance. Another side of detection with CNN, regression models are also applied to detection \cite{MASKCNN, OVERFEAT, DEEPMULTIBOX}, but direct mapping from an image to an exact bounding box is relatively difficult for a CNN. We thought that there must be a room for modification and improvement for the use of CNN as a regressor.

In this paper, we introduce a novel and straightforward detection method by integrating object existence estimation and bounding box optimization into a single convolutional network. Our model is on a similar line of a detection-by-regression approach, but we adopt a successful CNN-classification model rather than a CNN-regression model.
We estimate an exact bounding box by aggregating many weak predictions from the classification model, such as an ensemble method combines many weak learners to produce a strong learner.
We modify a traditional classification model to a suitable form, named \mbox{\textbf{\textit{AttentionNet}}}, for estimating an exact bounding box. This network provides quantized directions pointing a target object from the top-left (TL) and bottom-right (BR) corner of an image.

\Fref{FIG_TEASER} shows real detection examples of our method. Starting from an entire image, the image is recursively cropped according to the predicted directions at TL/BR and fed to \textit{AttentionNet} again, until the network converges to a bounding box fitting a target object. \textit{Each direction is weak but the ensemble of directions is sufficient to estimate an exact bounding box.} The difficulty of estimating an exact bounding box at once with a regression model is solved by a classification model. For multiple instances, we also place our framework on the sliding window paradigm but introduce an efficient method to cope with those.

Compared with the previous state-of-the-art methods, a single AttentionNet does everything but yields state-of-the-art detection performance. Our framework does not involve any separated models for object proposals nor post bounding box regression. A single AttentionNet 1) detects regions where a single instance is included only, 2) provides directions to an instance at each region, 3) and also correct the miss-localizations such as bounding box regression \cite{DPM}.

AttentionNet we demonstrate in this paper is not scalable to multiple classes yet.
Extension of this method to multiple classes with a single AttentionNet is ongoing.
We therefore demonstrate single-class object detection on public datasets to verify the strength of our model.
We primarily evaluate our detection framework by the human detection task and extend to a non-human class.

\paragraph{Contributions} Our contributions are three folds:
\begin{enumerate}
\item We suggest a novel detection method, which estimates an exact bounding box by aggregating weak predictions from AttentionNet.
\item Our method does not include any separated models such as the object proposal, object classifiers and post bounding box regressor. AttentionNet does all these.
\item We achieve the state-of-the-art performance on single-class object detection tasks.
\end{enumerate}

\begin{figure*}[t]
\begin{center}
\includegraphics[width=1\linewidth]{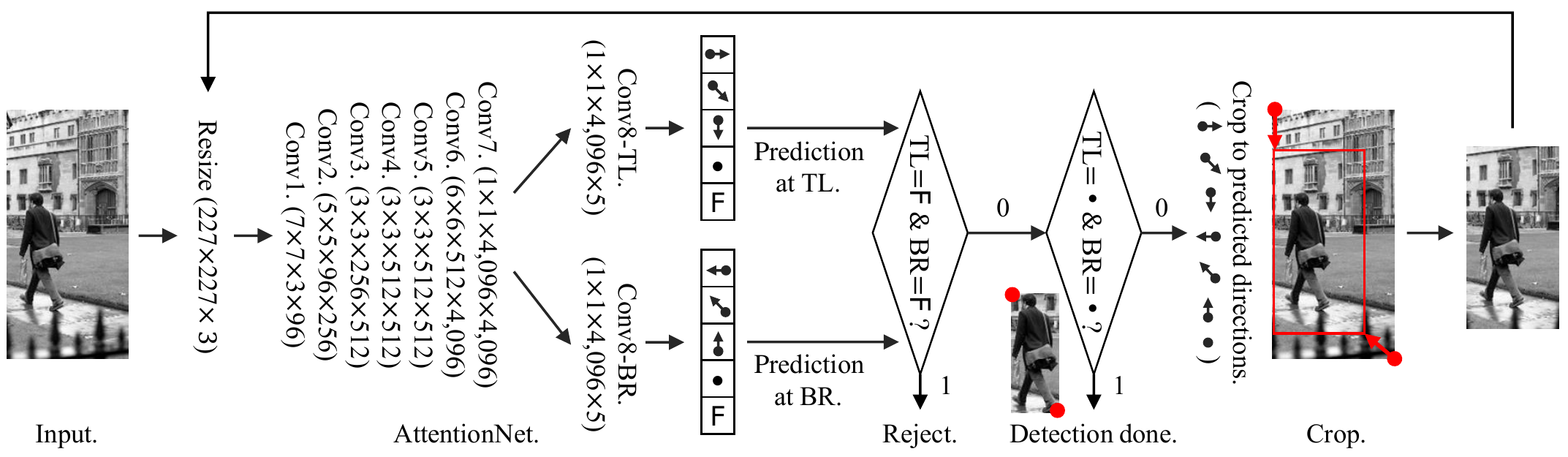}
\end{center}
   \caption{A pipeline of our detection framework. AttentionNet is composed of two final layers for top-left (TL) and bottom-right (BR) of the input image domain. Each of them outputs a direction ($\rightarrow$ $\searrow$ $\downarrow$ for TL, $\leftarrow$ $\nwarrow$ $\uparrow$ for BR) where each corner of the image should go to for the next step, or a ``stop'' sign ($\bullet$), or ``non-human'' sign (F). When AttentionNet outputs ``non-human'' in both layers, the image is rejected. The image is cropped according to the weak directions and fed to AttentionNet again, until it meets ``stop'' in both layers.}
\label{FIG_PIPELINE}
\end{figure*}

\section{Related Works}
Object detection has been actively studied for the last few decades.
One of the most popular approaches is part-based models due to their strength in handling pose variations and occlusions. Deformable Part Model (DPM), proposed by Felzenszwalb~\etal~\cite{DPM}, is a flexible model composed of object parts combined with deformation cost to manage severe deformations. Poselets, proposed by Bourdev~\etal~\cite{POSELETS}, is another part-based model demonstrating competitive performance. Poselets has numerous part-wise HOG detectors covering wide pose variations. Activated parts vote the location of objects.

In recent years, CNN-based approach leads to the successful development of object detection with drastic advances of deep CNN~\cite{LECUN}. Large scale ImageNet~\cite{IMAGENET} database and the raise of parallel computing contribute to a breakthrough in detection~\cite{MASKCNN, OVERFEAT, RCNN, DEEPIDNET, VERYDEEP, GOOGLENET} as well as classification~\cite{ALEX}. 

The state-of-the-art method in object detection is the Region-CNN (R-CNN), which represents a local region by CNN activations \cite{RCNN}. Specifically, R-CNN framework proceeds as follows. First it extracts local regions which probably contain an object by using an object proposal methods~\cite{SS}. The local regions, called object proposals, are warped into a fixed size and fed to a pre-trained CNN. Then each proposal is represented by mid-level CNN activations (e.g. 7th convolutional layer) and evaluated by separated classifiers (e.g. SVMs). The object proposals are then merged and fed to a bounding box regressor~\cite{DPM} to correct miss-localizations. Despite its efficiency and successful performance, it has a limitation that proposal quality highly affects detection performance. If the proposal model fails to propose a suitable candidate, the rest procedures will not have the opportunity to detect it. For this reason, \cite{DEEPMULTIBOX, DEEPMULTIBOX2} proposed a new class-agnostic proposal method with a CNN regression model to improve the proposal quality while reducing the number of proposals. Also, R-CNN is a cascaded model composed of individual components as object proposal, feature extraction, classification, and bounding box regression, therefore these should be individually engineered for the best performance.

Apart from R-CNN framework, there is another CNN-based approach, which considers object detection as a regression problem. Szegedy~\etal~\cite{MASKCNN} trains a CNN which maps an image to a rectangular mask of an object. Sermanet~\etal~\cite{OVERFEAT} also employ a similar approach but their CNN directly estimates bounding box coordinates. 
These methods are free from object proposals, but it is a still debatable to leave all to a CNN trained with a mean-square cost to produce an exact bounding box. 

Compared to the previous CNN methods, our method does not rely on object proposals since we actively explorer objects by iterative classifications. Also, the proposed network has a unified classification architecture, which is verified from many CNN applications and also does not need to tune up individual components.

\section{Detection with AttentionNet}
We introduce how our detection framework operates under a constrained condition where an image includes a single target instance only. Extension of this detection system to multiple instances will be described in \Sref{SEC_MULTI_INST}.
\subsection{Overview}

We summarize the core algorithm of our detection framework in \Fref{FIG_PIPELINE}. We warp an input image into a fixed CNN input size and feed it to AttentionNet. AttentionNet outputs two directional predictions corresponding the top-left (TL) corner and the bottom-right (BR) corner of the input image. For example, possible predictions for TL are the following: ``go to right ($\rightarrow$)'', ``go to right-down ($\searrow$)'', ``go to down ($\downarrow$)'', ``stop here ($\bullet$)'' and ``no instance in this image (F)''. Let us assume the prediction of AttentionNet indicates $\downarrow_{\text{TL}}$ and $\nwarrow_{\text{BR}}$ as shown in \Fref{FIG_PIPELINE}. We then crop the input image to the corresponding directional vectors of a fixed length $l$, and feed the cropped image to AttentionNet again to get next directions. This procedure is repeated until the image meets the following conditions: F in both corners, or $\bullet$ in both corners. If AttentionNet returns F at both corners, the test ends with a final decision of ``no target instance in this image''. Also, if AttentionNet returns $\bullet$ at both corners, the test ends with the current image as a detection result. This detected image can be back-projected to a corresponding bounding box in the original input image domain. Given a stopped (detected) bounding box $b$ and corresponding output activations $\mathbf{y}_{\text{TL}}, \mathbf{y}_{\text{BR}}\in \Re^5$, the detection score $s^b$ is discriminatively defined as,
\begin{equation}
\begin{split}
s^b&=s^b_{\text{TL}}+s^b_{\text{BR}},\quad \text{s.t.}\\
s^b_{\text{TL}}&=y^{\bullet}_{\text{TL}}-(y^{\rightarrow}_{\text{TL}}+y^{\searrow}_{\text{TL}}+y^{\downarrow}_{\text{TL}}+y^{\text{F}}_{\text{TL}}),\\
s^b_{\text{BR}}&=y^{\bullet}_{\text{BR}}-(y^{\leftarrow}_{\text{BR}}+y^{\nwarrow}_{\text{BR}}+y^{\uparrow}_{\text{BR}}+y^{\text{F}}_{\text{BR}}).
\label{EQ_SCORE}
\end{split}
\end{equation}

This detection framework has several benefits. Most portion of object proposal windows generated by traditional proposal methods truncate target objects. However, previous methods depends on these object proposals \cite{SSFISHER1, SSFISHER2, RCNN, DEEPIDNET, VERYDEEP} grade these windows by SVM scores only, without an intrinsic model to carefully handle the problem. A maximally scored object proposal through SVM does not guarantee that it is fitting an entire target object. This issue will be discussed in \Sref{SEC_VERIFICATION} again. In contrast, starting from a boundary out of target object, our method reaches at a terminal point with obvious stop signals in both corners like the examples in \Fref{FIG_TEASER}. Compared with previous methods solving the detection problem by a CNN-based regression \cite{OVERFEAT, MASKCNN}, our method adopts a more prosperous classification model with a soft-max cost rather than a regression model with a mean-square cost. Even if the model provides weak direction vectors which are quantized and have a fixed short length, our prediction for object detection becomes stronger as the bounding box is iteratively narrowed down to a target object.

\begin{figure}[t]
\begin{center}
\small
\includegraphics[width=1\linewidth]{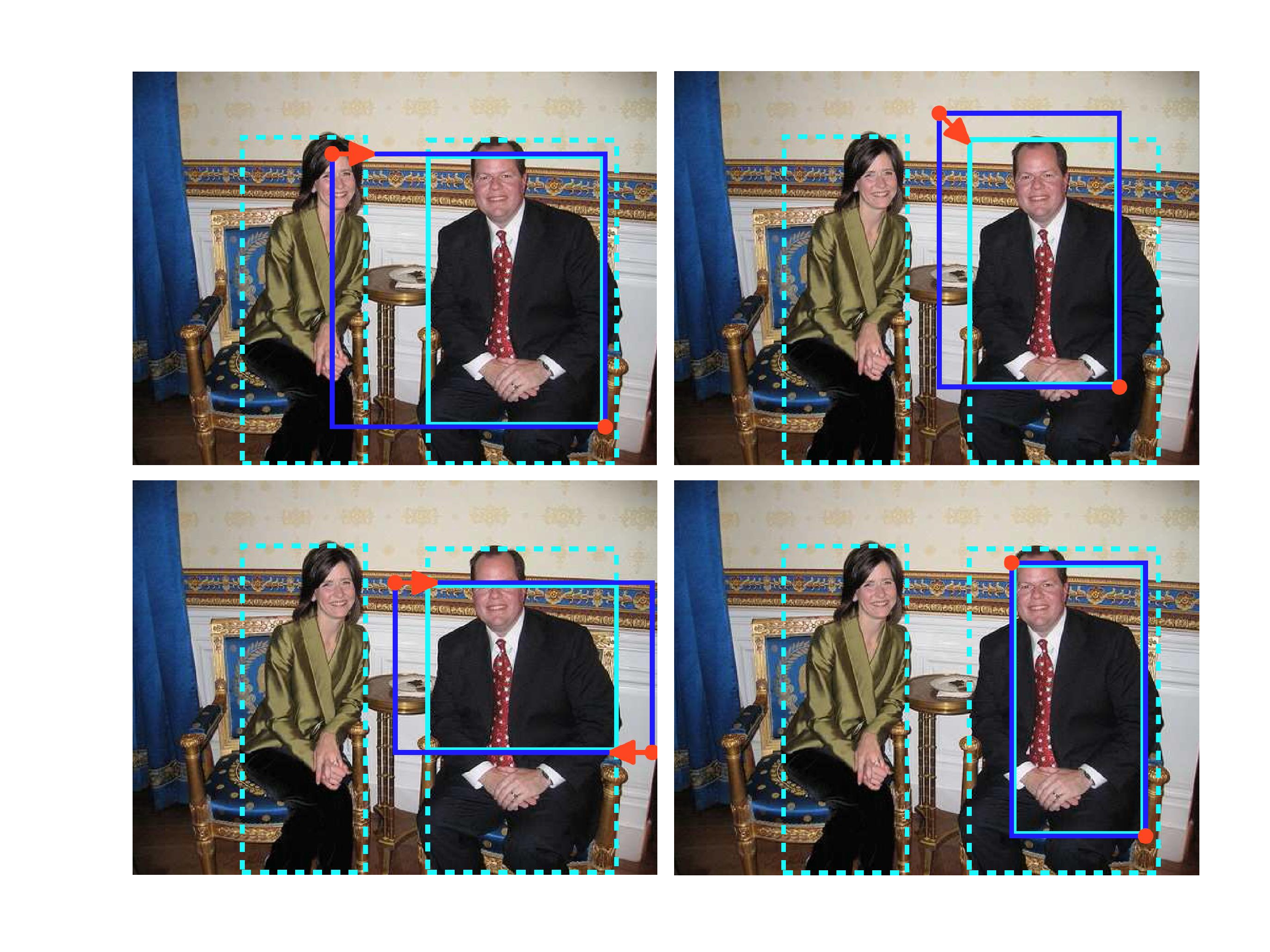}
\end{center}
\caption{Real examples of crop-augmentation for training AttentionNet. The target instance is the right man. Dashed cyan bounding boxes are ground-truths, and the blue bounding boxes are the augmented regions. Red arrows/dots denote their ground truths.}
\label{FIG_TRAINING_SAMPLES}
\end{figure}

\subsection{Network architecture}
The architecture of AttentionNet is illustrated in \Fref{FIG_PIPELINE} but we drop pooling and normalization layers from the figure for simplicity. Layers from the first to the seventh convolution (also called seventh fully-connected layer) are the same layers in Chatfield \etal's VGG-M network \cite{DEVIL}. In this network, the stride and filter size are smaller at the first layer but the stride at the second convolution layer is larger, compared with Krizhevsky \etal's network \cite{ALEX}. This is also similar to the CNN suggested by Zeiler and Fergus \cite{MATTHEW}. We adopt this model due to its superior performance on the ILSVRC classification (Top-5 error of 16.1 on a single model) without significant increase in training time. Please refer to \cite{DEVIL} for more details of this architecture.

Our detection framework requires two predictions for TL and BR, but we prefer a CNN for classification to be trained with soft-max loss. While the regression with a mean-squre loss forces the model to a difficult bounding box, classification of quantized directions with soft-max loss forces the model to be maximally activated at a true direction rather than exact values of a bounding box coordinates. We therefore separate the output layers of TL and BR, because this setting enables us to use the soft-max loss with which each branch returns each prediction. These final layers are denoted by Conv8-TL and Conv8-BR in \Fref{FIG_PIPELINE}. Confidences of the 5 decisions at TL (BR) are computed with corresponding 5 filters of 1$\times$1$\times$4,096 size at Conv8-TL (Conv8-BR). These two layers are followed by the final ReLU layer.

\subsection{Training}
To make AttentionNet operate in the scenario we devise, it is quite important to process original training images to a suitable form. During the test stage, starting from an initial test over the entire image boundary to a final decision of ``stop'' or ``no instance'', the number of possible decision pairs is 17 ($=$4$\times$4$+$1) such as $\left\{\rightarrow,\searrow,\downarrow,\bullet\right\}_{\text{TL}}\times\left\{\leftarrow,\nwarrow,\uparrow,\bullet\right\}_{\text{BR}}$ for positive regions and $\left\{\text{F}_{\text{TL}},\text{F}_{\text{BR}}\right\}$ for negative regions. We therefore must augment the original training images into a reformed training set evenly covering these 17 cases. \Fref{FIG_TRAINING_SAMPLES} shows real examples how we process an original training image to multiple augmented regions. We randomly generate positive regions which satisfy the following three rules.
\begin{enumerate}
\item A positive region must include at least 50\% of the area of a target instance.
\item A positive region can include multiple instances (as the top-left example in \Fref{FIG_TRAINING_SAMPLES}), but the target instance must occupy the biggest area. Within a cropped region, the area of the target instance must be at least 1.5-times larger than that of the other instances.
\item Regions are cropped in varying aspect ratios as well as varying scales.
\end{enumerate}
The second rule is important for complex instance layouts in the multiple instance scenario (to be introduced in \Sref{SEC_MULTI_INST}). Without this rule in the scenario, a final bounding box is prone to fit multiple instances at once. In order to make AttentionNet always narrow the bounding box down to the largest instances among multiple instances, we must follow the second rule in generating positive regions. The third rule is also necessary because aspect ratio and scale change during the iterative crops in a test stage. We extract negative regions which are not overlapped with bounding boxes of a target class, or overlapped with bounding boxes of non-target classes. The negative regions also have varying aspect ratios and scales.

\begin{figure*}[t]
\begin{center}
\includegraphics[width=1\linewidth]{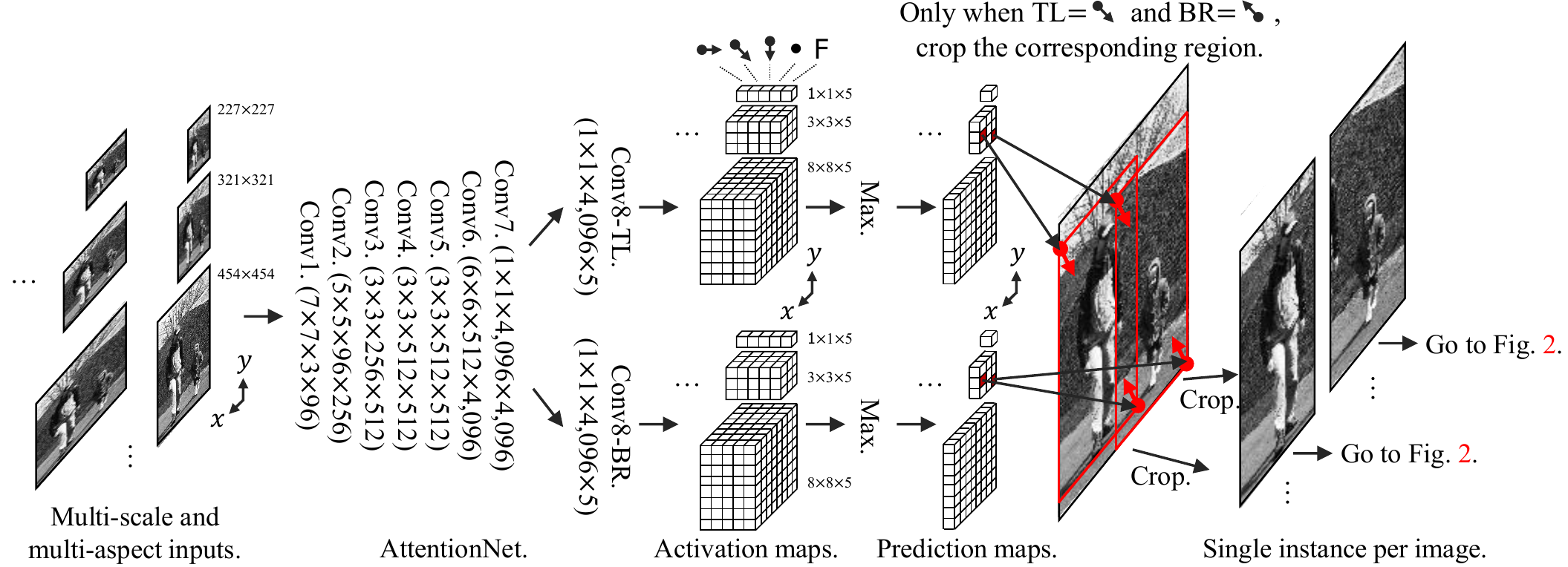}
\end{center}
   \caption{Extracting single-instance regions, where a single instance is included only. Multiple inputs with multiple scales/aspects are fed to AttentionNet, and prediction maps are produced. Only image regions satisfying $\left\{\searrow_{\text{TL}},\nwarrow_{\text{BR}}\right\}$ are regarded as the single-instance regions. These regions are fed to AttentionNet again for final detection. Note, the CNN here and that in \Fref{FIG_PIPELINE} are the same one, not separated.}
\label{FIG_MULTI_INST}
\end{figure*}

When we compose a batch to train the CNN, we select positive and negative regions in an equal portion. In a batch, each of the 16(=4$\times$4) cases for positive regions occupies a portion of 1/(2$\times$16), and the negative regions occupy the remaining portion of 1/2. The loss for training AttentionNet is an average of the two soft-max losses computed independently in TL and BR.

\subsection{Verification of AttentionNet}
\label{SEC_VERIFICATION}
Before we extends AttentionNet to multiple-instances, we verify the effectiveness of our top-down approach, against to the object proposal \cite{SS} based framework \cite{RCNN}. 
As studied by Agrawal \etal \cite{ANALYZE_CNN}, strong mid-level activations in a CNN come from object ``parts'' that is distinctive to other object classes. Because Region-CNN based detection relies on each region score coming from the CNN activations, it is prone to focus on discriminative object ``part'' (e.g. face) rather than ``entire object'' (e.g. entire body).

To analyze this issue, we demonstrate a toy experiment of human detection. 
As an object proposal based method, we represent object proposals by activations from a fine-tuned CNN and score them by a SVM. 
Then, we choose the bounding box which has the maximum SVM score as a detection result. 
We compare this setting with our AttentionNet framework in \Fref{FIG_PIPELINE}. 
The Region-CNN, SVM, and AttentionNet are trained with the same data including ILSVRC'12 classification and PASCAL  VOC 2007. The test images which contain a single human with a reasonable scale are selected from PASCAL VOC 2007 testset. Then, we compute the average precision.

\begin{figure}[t]
	\begin{center}
		\includegraphics[width=1\linewidth]{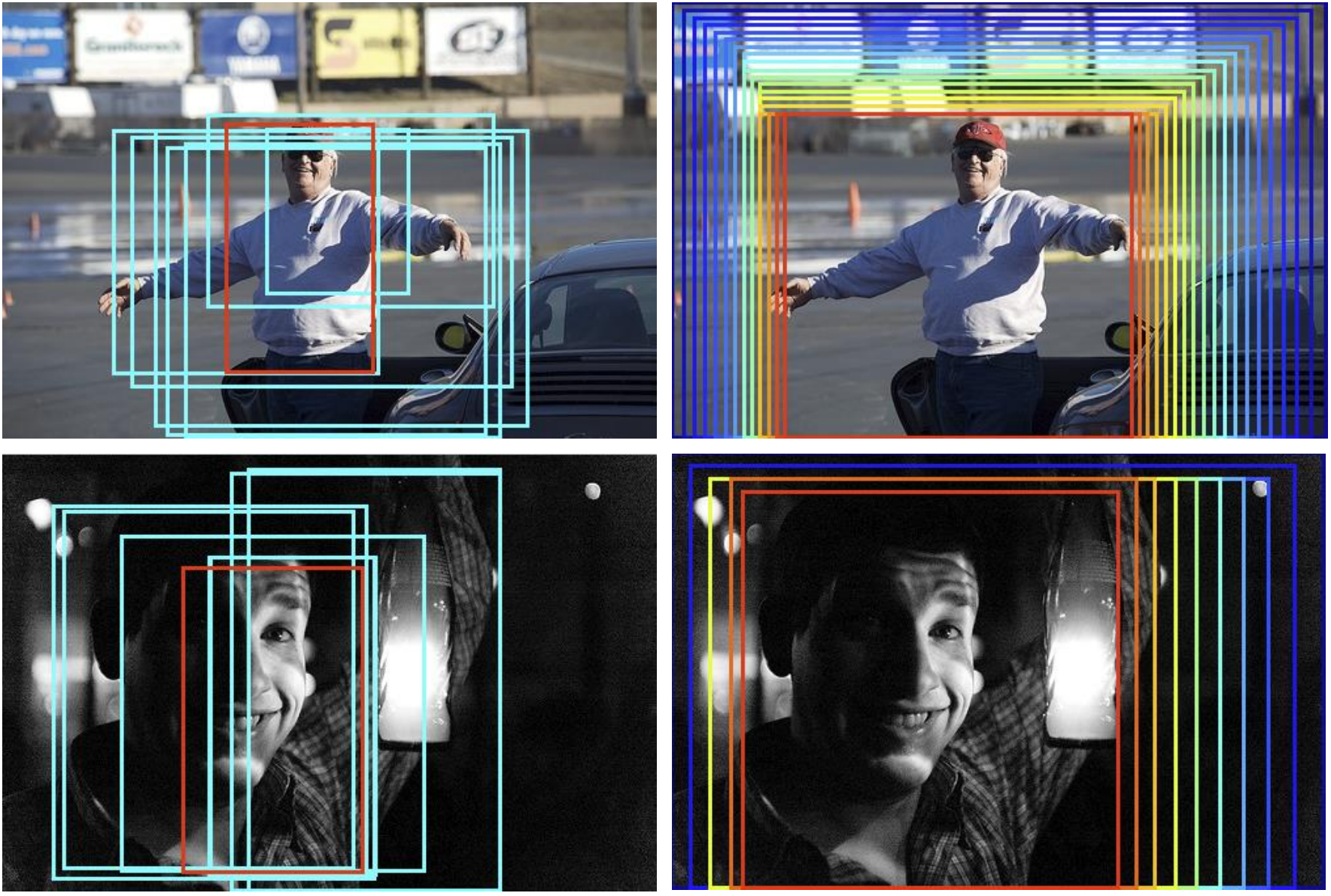}
	\end{center}
	\caption{Real detection examples of the object proposal based method (left) and AttentionNet (right). In the left column, a red bounding box is the top-1 detected region among top-10 object proposals (cyan) with the maximum SVM score.}
	\label{FIG_TOY_EXP}
\end{figure}

In this experiment, the object proposal based setting shows 79.4\% while AttentionNet shows 89.5\%. The object proposal based method shows much lower detection performance because of the relatively weak correlation between strong activation and ``entire'' human body. In contrast, AttentionNet reaches a terminal point well starting from a boundary out of a target object. As shown in \Fref{FIG_TOY_EXP}, the maximally scored object proposal is prone to focus on discriminative faces rather than ``entire'' human body.


\section{Extension to Multiple Instances}
\label{SEC_MULTI_INST}
AttentionNet is trained to detect a single instance in an image and operates in this way. In this section, we introduce an efficient method to extend our detection framework to a practical situation where an image involves multiple instances. Our straightforward solution is to propose candidate regions where only a single instance is included. We reuse AttentionNet for the single instance region proposal, therefore no separated model is necessary. We then detect each instance from each region proposal, and merge the results into a reduced number of bounding boxes followed by a final refinement procedure where AttentionNet is also reused again.

\begin{figure*}
	\begin{center}
		\small
		\begin{tabular}{ccccc}
			\setlength{\tabcolsep}{1.7pt}
			\includegraphics[width=0.17\linewidth]{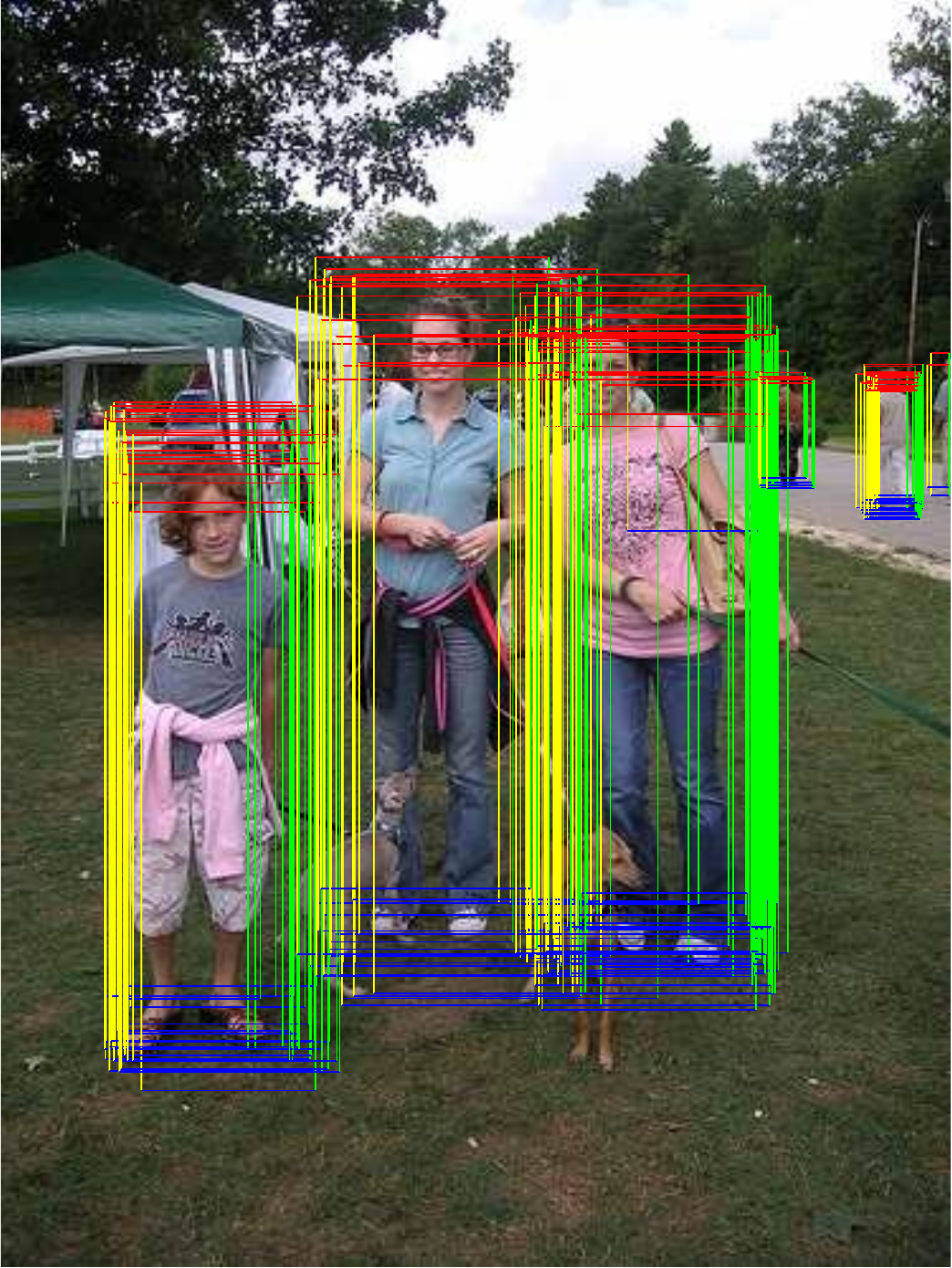}&
			\includegraphics[width=0.17\linewidth]{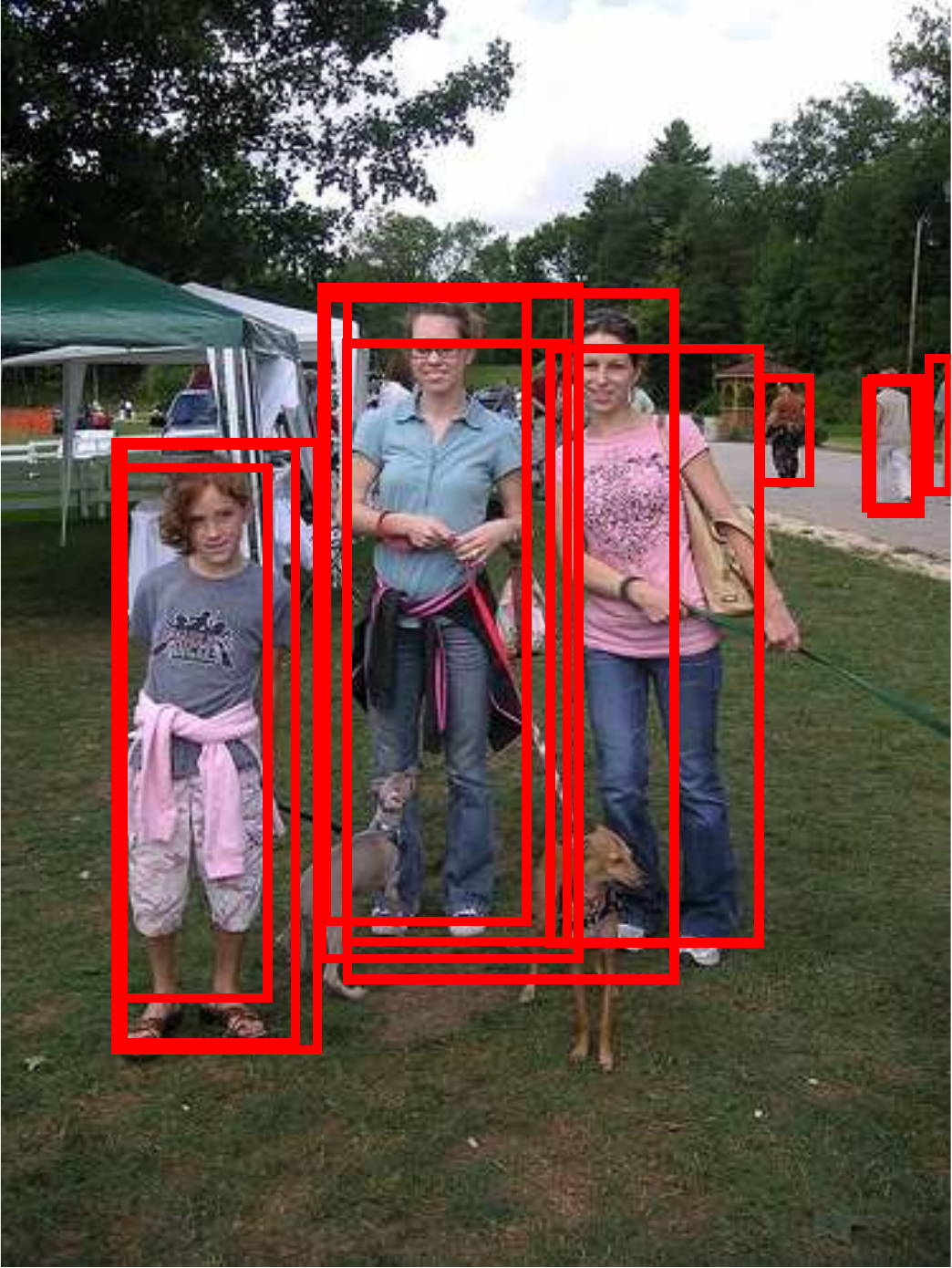}&
			\includegraphics[width=0.17\linewidth]{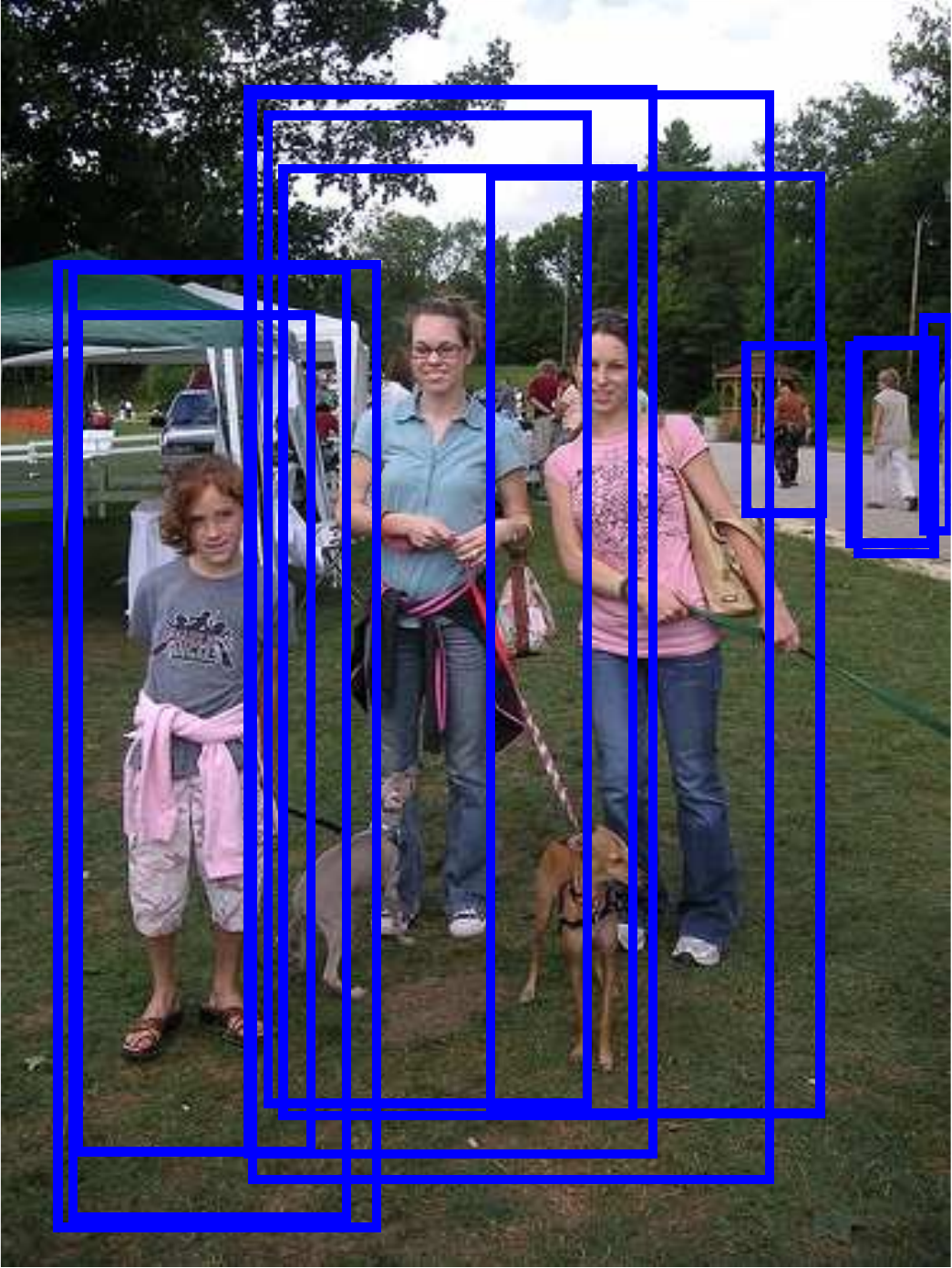}&
			\includegraphics[width=0.17\linewidth]{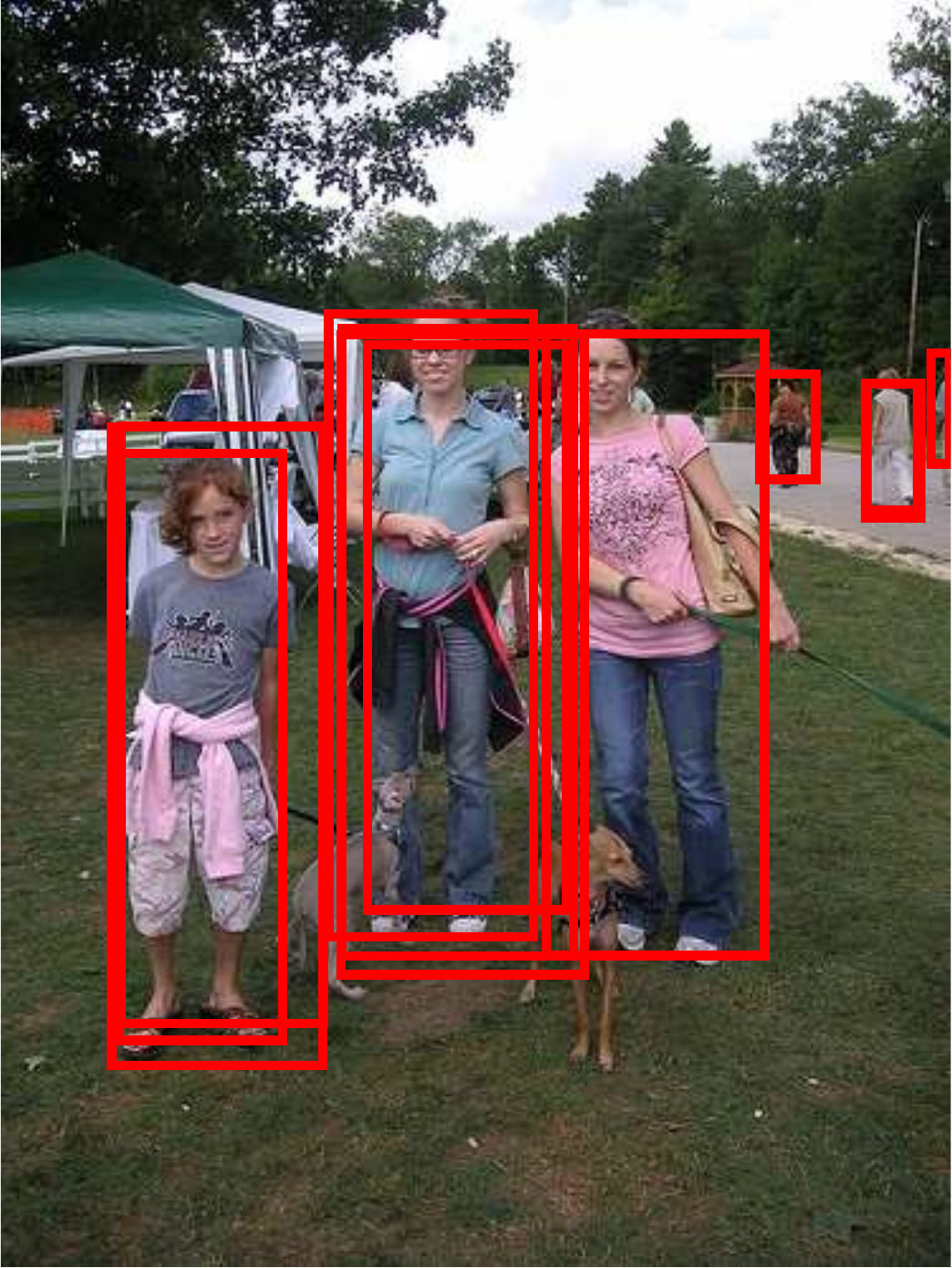}&
			\includegraphics[width=0.17\linewidth]{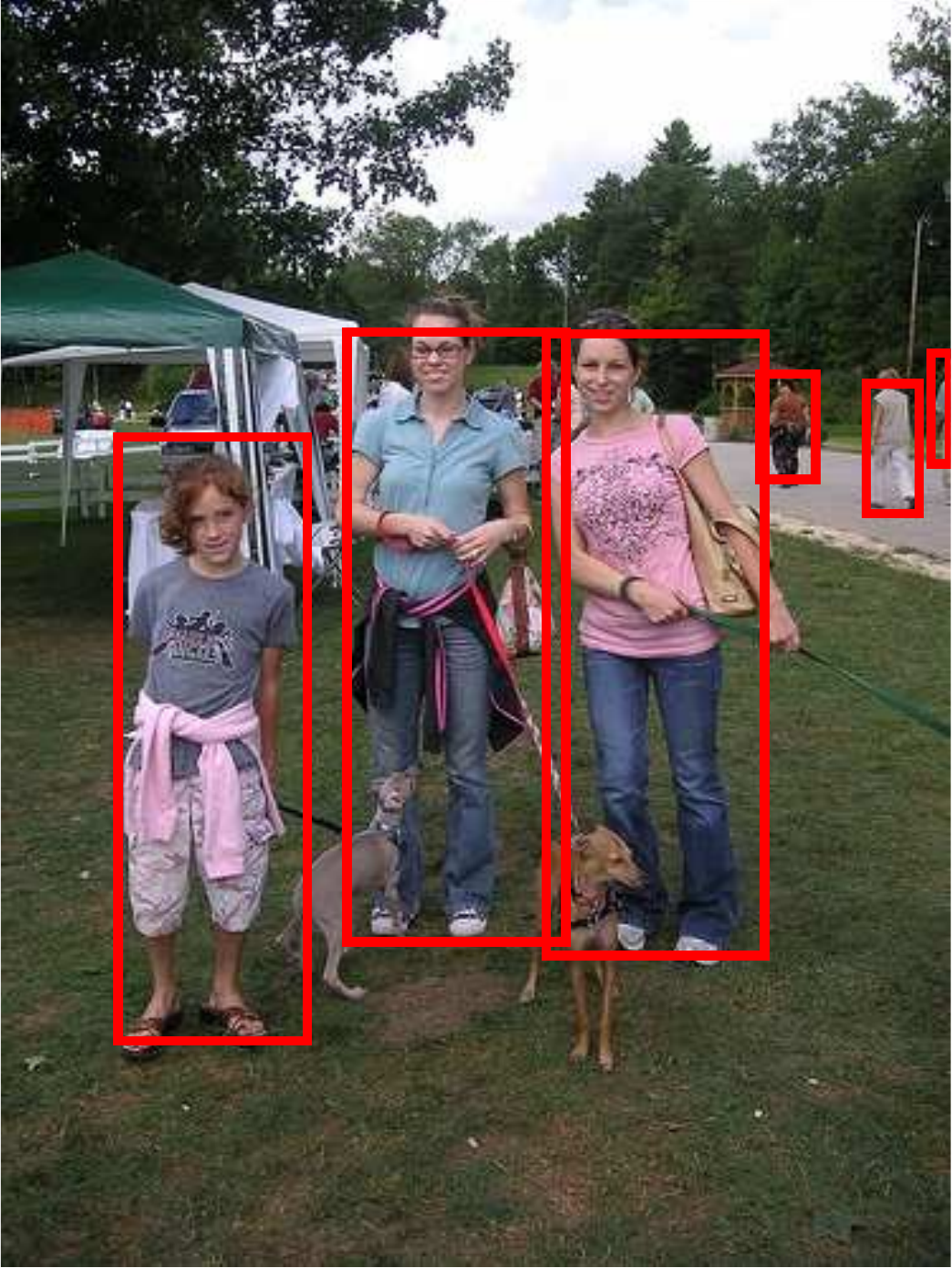}\\
			(a) Initial detections.&(b) Initial merge.&(c) Re-initialize. ($\times$2.5)&(d) Re-detections.&(e) Final merge.
		\end{tabular}
	\end{center}
	\caption{Real examples of our detection procedure, including initial results (a$\sim$b) and refinement (c$\sim$e). Initially detected candidates come from \Fref{FIG_MULTI_INST} followed by \Fref{FIG_PIPELINE} are merged by an intersection over union (IoU) of 0.8. We extend each merged box to 2.5-times larger size, and feed them to \Fref{FIG_PIPELINE} again. Finally we merge the second results by an IoU of 0.5.}
	\vspace{-2mm}
	\label{FIG_DET_PROCEDURE}
\end{figure*}

\subsection{Efficient single instance region proposal}
\label{SEC_REGION_PROPOSAL}

Let us assume we have an arbitrary region in an image. If we feed this region to AttentionNet, the 17 decision combinations in TL/BR are possible. Among them, only the output of $\left\{\searrow_{\text{TL}},\nwarrow_{\text{BR}}\right\}$ guarantees that the region includes entire body of a target instance with proper margins. In the other decision combinations, the region is probably truncating a target instance or does not contain an instance. This is the logic to make sure whether an entire single instance is included or not.

To boost the recall of the single instance region proposals, we place our framework in the sliding window paradigm, but we do not naively crop and feed each of sliding windows to AttentionNet. Following a technique successfully used in \cite{CNNPOSE, YOO}, we also utilize the property that a CNN does not require fixed size input only because a fully connected layer is a convolution layer composed of filters of 1$\times$1 size. For example, if an image 2-times larger (321$\times$321) than a regular CNN input (227$\times$227) is fed to AttentionNet, a spatial activation map of 3$\times$3$\times$5 size is produced from an output layer as shown in \Fref{FIG_MULTI_INST}. Here, the 9 activation vectors ($\in\Re^5$) correspond to 9 local patches of 227$\times$227 size with a stride of 32. In this way, we can speed up the sliding window method with AttentionNet.

Since instances in images are possibly diverse in aspect ratio and scale, multiple scales/aspects are also required for the sliding windows. We therefore feed multi-scale/aspect images to AttentionNet, and obtain predictions of thousands of sliding windows, as depicted in \Fref{FIG_MULTI_INST}. Given the prediction maps, only corresponding regions of the prediction of $\left\{\searrow_{\text{TL}},\nwarrow_{\text{BR}}\right\}$ are cropped and fed to AttentionNet again for the final detection. The number of scales/aspects can be deterministically set according to the size/aspect statistics of ground-truth bounding boxes in a training set.

\subsection{Initial detection and final refinement}

Each single instance region proposal produced in \Sref{SEC_REGION_PROPOSAL} is fed to AttentionNet iteratively until it meets $\left\{\bullet_{\text{TL}},\bullet_{\text{BR}}\right\}$ or $\left\{\text{F}_{\text{TL}},\text{F}_{\text{BR}}\right\}$. The first image in \Fref{FIG_DET_PROCEDURE} shows a real example of the initial detections. These bounding boxes are merged to a decreased number by single-linkage clustering: a group of bounding boxes satisfying a minimum intersection over union (IoU) of $\alpha_{0}$ are averaged into one with their scores of \Eref{EQ_SCORE}.

To refine the result, \cite{DPM, RCNN} employ a bounding box regression, which finally re-localizes the bounding boxes. This is a linear regression model which maps a given feature of a bound box to a new detection window. 

In our case, we can employ AttentionNet again as the role for bounding box regression for further refinement. We re-scale each bounding box in \Fref{FIG_DET_PROCEDURE}-(b) to a new initial window in \Fref{FIG_DET_PROCEDURE}-(c) by a re-scaling factor of $\beta$. These re-initialized windows are fed to AttentionNet again and result in new bounding boxes as shown \Fref{FIG_DET_PROCEDURE}-(d). This re-detection procedure gives us one more chance to reject false positives as well as fine localization. These bounding boxes are finally merged to final results by an IoU of $\alpha_{1}$. 

\section{Evaluation}
We perform the object detection task on public datasets to verify the strength of AttentionNet. We primarily apply our detection framework to a human detection problem and extend it to a non-human class. 

Among a wide range of object classes, it is beyond question that the class ``human'' has taken the center stage in object detection for decades because of its wide and attractive applications. Nonetheless, human detection on uncontrolled natural images is still challenging due to severe deformations caused by pose variation, occlusion, and overlapped layout. For rigorous verification in such challenging settings, we choose human as our primary target class.

\paragraph{Datasets}
We select PASCAL VOC 2007 and 2012 \cite{PASCAL} where images are completely uncontrolled because they are composed of arbitrary web images from Flickr. Humans in these sets are severely occluded, truncated and overlapped with diverse pose variations as well as scales. Following the standard setting in previous human detection over these sets, we use all the ``trainval'' images for training, and report an average precision (AP). For PASCAL VOC 2007, we rigorously use the evaluation function provided by the development toolkit. For PASCAL VOC 2012, we submit our result to the evaluation server and obtain the AP value. 
\paragraph{Training} 
Our framework requires only one training stage for AttentionNet. The selected datasets provide relatively small number of training images, while AttentionNet is composed of many weights to be optimized. Thus, we pre-train the model with ILSVRC 2012 classification dataset \cite{ILSVRC} and transfer the model to our target datasets. Following \cite{ANALYZE_CNN}, the learning rate of the pre-trained layers (e.g. 0.001 for conv1 to conv7) is 10-times smaller than that of the re-initialized layer (e.g. 0.01 for conv8-TL and conv8-BR).

\paragraph{Parameters in test stage}
Our method requires several parameters to be determined before the test stage. We tuned all these parameters over the validation set of PASCAL VOC 2007. After we determine the proper parameter values, we apply the same parameters to the whole experiments in this paper regardless of datasets. We use the parameters as follows. We set the length $l$ of each direction vector to 30 pixels, and we limit the maximum number of iterative feed-forwards to 50 to prevent the possibility of divergence. 
To increase the chance for detecting large instances (e.g. an image fully filled with a face), we first re-size the average image to ($2h\times2w$)-size and then place an input image of ($h\times w$)-size at the center of the magnified average image before feed-forwarding. We use 7 scales with a scale step of 2, and 3 aspect ratios of $\left\{\frac{h}{w}\;|\;1.0, 1.5, 2.0\right\}$, according to the statistics of ground-truth bounding boxes in the training set. We set the merging parameters of $\alpha_0$ and $\alpha_1$ to 0.8 and 0.5, respectively. We set the re-scaling factor $\beta$ in \Fref{FIG_DET_PROCEDURE}-(c) to 2.5. When we do not perform the refinement step of \Fref{FIG_DET_PROCEDURE}-(c$\sim$e), we set the initial merging parameter $\alpha_0$ to 0.6.

We first evaluate the performance of human detection on PASCAL VOC 2007/2012. We compare our method against the recent state-of-the-art methods, and \Tref{TAB_VOC2007} shows the result. Without the refinement step of \Fref{FIG_DET_PROCEDURE}-(c$\sim$e), our method achieves 61.7\% and 62.8\% for each dataset. When we equip the refinement step, we achieve a new state-of-the-art score of 65.0\% and 65.6\% with an additional improvement of +3.3\% and +2.8\%. For the refinement step, we simply re-use AttentionNet for re-localization, therefore we do not need an extra model. 

\begin{table}
\setlength{\tabcolsep}{1.0pt}
\small
\begin{center}
\begin{tabular}{|l|l|c|c|}
\hline
Method&Extra data&VOC'07&VOC'12\\
\hline\hline
AttentionNet&ImNet&61.7&62.8\\
AttentionNet + Refine&ImNet&\textbf{65.0}&\textbf{65.6}\\
\hline\hline
AttentionNet + R-CNN&ImNet&66.4&69.0\\
AttentionNet + Refine + R-CNN&ImNet&\textbf{69.8}&\textbf{72.0}\\
\hline\hline
Person R-CNN + BBReg&ImNet&59.7&N/A\\
Person R-CNN + BBReg$\times 2$&ImNet&\textbf{59.8}&N/A\\
Person R-CNN + BBReg$\times 3$&ImNet&59.7&N/A\\
\hline\hline
Felzenszwalb \etal'10 \cite{DPM}&None.&41.9&N/A\\
Bourdev \etal'10 \cite{POSELETS}&H3D&46.9&N/A\\
Szegedy \etal'13 \cite{MASKCNN}&VOC'12&26.2&N/A\\
Erhan \etal'14 \cite{DEEPMULTIBOX}&None.&37.5&N/A\\
Gkioxari \etal'14 \cite{POSELETS_KEYPOINT}&VOC'12&45.6&N/A\\
Bourdev \etal'14 \cite{DEEPPOSELETS}&ImNet + H3D&59.3&58.7\\
He \etal'14 \cite{SPP}&ImNet&57.6&N/A\\
Girshick \etal'14 \cite{RCNN}&ImNet&58.7&57.8\\
Girshick \etal'14 \cite{RCNN}&ImNet&\textbf{64.2*}&N/A\\
Shen and Xue '14 \cite{SHEN}&ImNet&59.1&\textbf{60.2}\\
\hline
\end{tabular}
\\\textbf{*}\footnotesize Very deep model of 16 convolution layers \cite{VERYDEEP} is used.
\end{center}
\caption{Human detection performance on PASCAL VOC 2007/2012. ImNet denotes ILSVRC 2012 classification set. AP(\%) is reported.}
\label{TAB_VOC2007}
\end{table}

Similar to ours, Poselets-based methods \cite{POSELETS, POSELETS_KEYPOINT, DEEPPOSELETS} are also limited to a single-class (e.g. human) object detection. Our method outperforms these methods with a large margin. Though we does not include an intrinsic human model to handle diverse poses and severe occlusions, our framework successfully converged to the window fitting the human from a window coarsely covering an entire human body. Our top-down approach is robust to the diverse human poses (\Fref{FIG_TEASER}) because our model is trained to operate in that way with carefully augmented training samples. 

In the detection-by-regression manner, most similar works to ours are \cite{MASKCNN, DEEPMULTIBOX}. CNN-regression models combined with a mean square error are trained to produce a target object mask \cite{MASKCNN}, or bounding box coordinates \cite{DEEPMULTIBOX} for the purpose of class-agnostic object proposals. Our AttentionNet clearly outperforms these methods, and it verifies the strength of the ensemble of weak directions from a promising CNN-classification model.

R-CNN \cite{RCNN} has been the best performing detection framework which is used for \cite{SPP, SHEN} showing state-of-the-art performances, but our method also beats these methods which use the 8-layered architecture. Our result is even slightly better (+0.8\%) than that of R-CNN \cite{RCNN} equipped with a 16-layered very deep network~\cite{DEVIL}, which yields the top-5 error of 7.0\% in ILSVRC'12 classification. However, one obvious strength of R-CNN is class-scalability while yielding high performance. Extension to multiple classes with a single AttentionNet is our primary direction.

To make the comparison completely fair, we trained R-CNN on the setting of ``person''-versus-all by using the R-CNN code\footnote{https://github.com/rbgirshick/rcnn} provided by the authors. It is noted by ``Person R-CNN + BBReg''. Our method still shows better results, demonstrating the significant margin of +5.3\% on PASCAL VOC 2007. We also tried to iterate the bounding-box regression in R-CNN. It is noted by ``BBReg$\times N$''. The improvement is negligible: +0.1\% and +0.0\% for second and third iterations. These imply the benefit of our stacked classification strategy against to the stacked regression.

\begin{figure}[t]
	\begin{center}
		\begin{tabular}{@{}c@{ }c@{}}
		\includegraphics[width=0.49\linewidth]{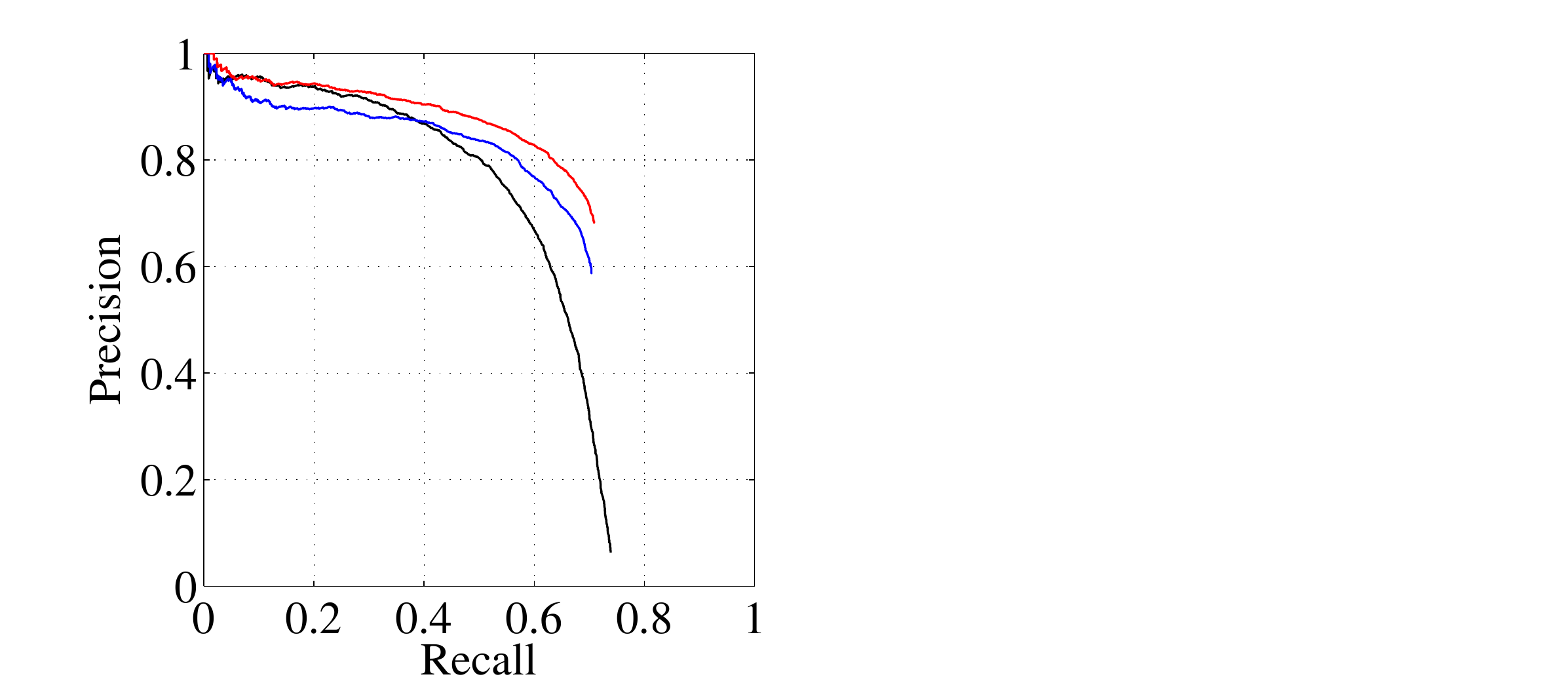} &
		\includegraphics[width=0.49\linewidth]{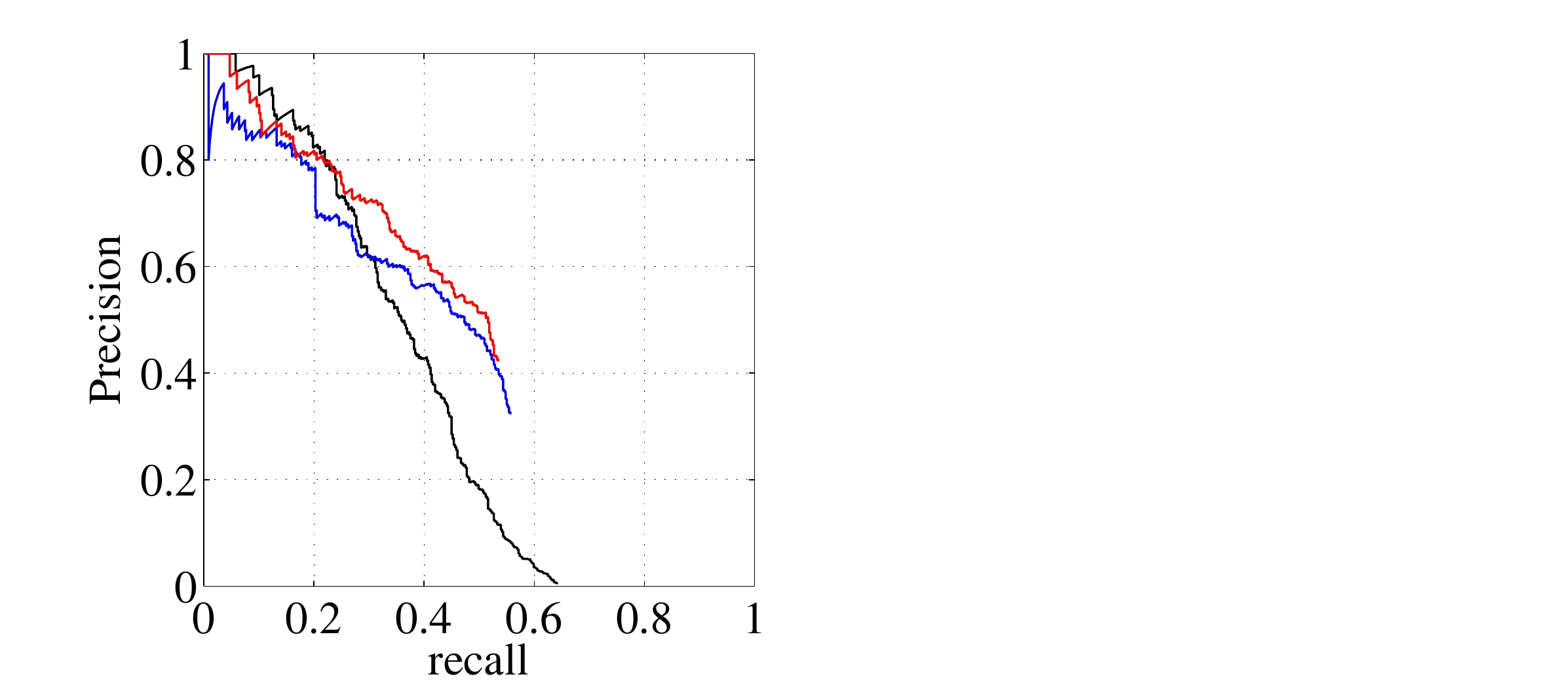} \\
		\hspace{15pt}\includegraphics[width=0.43\linewidth]{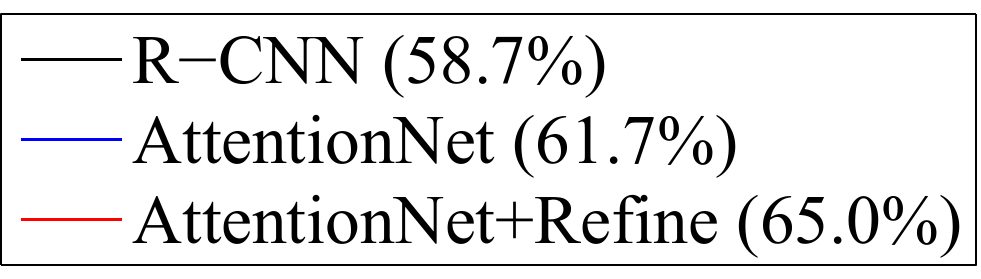} &
		\hspace{15pt}\includegraphics[width=0.43\linewidth]{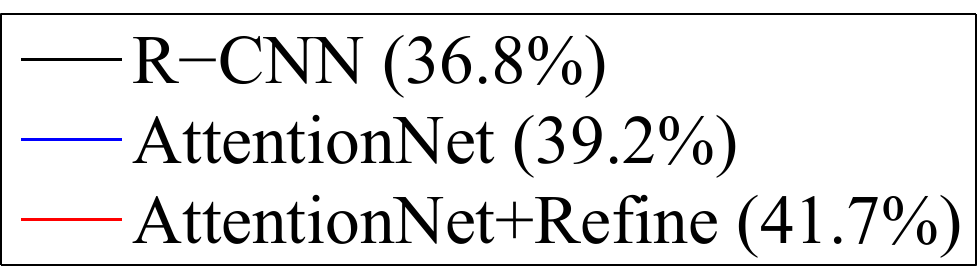} \\
		\small (a) Person class & \small (b) Bottle class
		\end{tabular}
	\end{center}
	\caption{Precision-recall curves on PASCAL VOC 2007.}
	\label{FIG_PR_PERSON_BOTTOLE}
\end{figure}

\Fref{FIG_PR_PERSON_BOTTOLE}-(a) shows the precision-recall curves of human detection on PASCAL VOC 2007. We plot the curve of R-CNN by using the source code from the authors. The curve clearly shows both of strength and weakness of AttentionNet. Compared with R-CNN, our method yields much high precision, but the tail is truncated early. It implies lower recall than R-CNN. This tendency comes from our hard decision strategy: AttentionNet 1) accepts a region as an initial candidate only if the both corners satisfy $\left\{\searrow_{\text{TL}},\nwarrow_{\text{BR}}\right\}$ at the same time, and 2) takes a final positive decision only if the both corners indicate ``stop''. This hard criteria gives us significantly reduced number of bounding boxes with a quite strong confidence, but results in weak recall. \textit{Our AP of 65.0\% is a result only with 4,863 boxes, while R-CNN yields 58.7\% with 53,624 boxes.} 

We can derive a positive meaning from this observation: detection results from the two different approaches are complementary. We therefore combine the two results as in the following. Because our bounding boxes are more confident, we rejects R-CNN bounding boxes which are overlapped to ours more than an IoU of 0.7, and add a bias to our bounding box scores to be followed by the remaining R-CNN bounding boxes. As reported in \Tref{TAB_VOC2007}, we achieve the significantly boosted performance of 69.8\% from the combination of AttentionNet and R-CNN.

One AttentionNet is currently handle only one object class yet, however its application does not limited to specific classes since it does not include any crafted class-specific model. We therefore demonstrate performance on another object class. Among the 20 classes in PASCAL VOC series, ``bottle'' is one of the most challenging object class. Bottle is the smallest class, occupying only 8k pixels in average, while the other classes occupy 38k pixels in average. Its appearance is also indistinct due to its transparent material. Thus, we select ``bottle'' and verify our method for this challenging case. We use the exactly same parameters used in human detection without any tuning.

\begin{table}
\setlength{\tabcolsep}{1.0pt}
\small
\begin{center}
\begin{tabular}{|l|l|c|c|}
\hline
Method&Extra data&VOC'07&VOC'12\\
\hline\hline
AttentionNet&ImNet&39.2&41.2\\
AttentionNet + Refine&ImNet&\textbf{41.7}&\textbf{42.5}\\
\hline\hline
AttentionNet + R-CNN&ImNet&42.9&43.2\\
AttentionNet + Refine + R-CNN&ImNet&\textbf{45.5}&\textbf{45.0}\\
\hline\hline
Bottle R-CNN + BBReg&ImNet&32.4&N/A\\
Bottle R-CNN + BBReg$\times 2$&ImNet&32.4&N/A\\
Bottle R-CNN + BBReg$\times 3$&ImNet&\textbf{32.5}&N/A\\
\hline\hline
He \etal'14 \cite{SPP}&ImNet&40.5&N/A\\
Girshick \etal'14 \cite{RCNN}&ImNet&36.8&32.6\\
Girshick \etal'14 \cite{RCNN}&ImNet&\textbf{44.6*}&N/A\\
Shen and Xue '14 \cite{SHEN}&ImNet&36.3&\textbf{33.1}\\
\hline
\end{tabular}
\\\textbf{*}\footnotesize Very deep model of 16 convolution layers \cite{VERYDEEP} is used.
\end{center}
\caption{Bottle detection performance on PASCAL VOC 2007/2012. AP(\%) is reported.}
\label{TAB_VOC2012_BOTTLE}
\end{table}

\Tref{TAB_VOC2012_BOTTLE} shows the result of bottle detection on PASCAL VOC 2007/2012. We observe a similar tendency to human detection in overall. Except R-CNN \cite{RCNN} equipped with very large CNN \cite{VERYDEEP}, our method yields the best scores in both datasets. Specifically, AttentionNet with refinement beats the previous methods by the gaps more than +9.4\% in VOC 2012. Our method combined with the complementary R-CNN still gives us performance improvements.

\Fref{FIG_PR_PERSON_BOTTOLE}-(b) shows the precision-recall curves of bottle detection on PASCAL VOC 2007. Although R-CNN shows slightly better precision in low recall, their precision steeply decreases while our precision decreases with a less slope. Our curve is truncated early with a lower recall than R-CNN. Further positive mining and bootstrapping could be possible solutions to boost recall.

\section{Conclusions}
In this paper, we have proposed a novel method for object detection. 
We adopted a well-studied classification technique into object detection and presented AttentionNet to predict weak directions towards a target object. Since we actively explorer an exact bounding box of a target object in a top-down approach, we does not suffer from the quality of initial object proposals and provide accurate object localization.

Through this work, we have two important observations. Firstly, we achieve the new state-of-the-art performance by changing the way of object detection. Second, our top-down approach is complementary to the previous state-of-the-art method using a bottom-up approach, therefore combining two approaches boosts the performance of object detection.


\paragraph{Limitations and future work}
We have two limitations in this study. 
Our AttentionNet is not scalable to multiple classes. However, it is a fact that AttentionNet has a potential for extension to generic object classes, because this model does not include any crafted class-specific models. The other thing is low recall. It is caused by our hard decision strategy as we stated. We believe there is a room for boosting recall. For example, we can loosen the hard decision criteria by employing a thresholding strategy to the direction scores from AttentionNet. Positive mining and bootstrapping can also be a promising candidate solutions. We also leave these issues as a future work.

\section{Acknowledgement}
This work was supported by the Technology Innovation Program (No. 10048320), funded by Korea government (MOTIE). This work was also supported by the National Research Foundation of Korea (No. 2010-0028680), funded by Korea government (MSIP).

{\small
\bibliographystyle{ieee}
\bibliography{egbib}
}

\end{document}